# A Formalization of the Turing Test

Evgeny Chutchev

## 1. Introduction

The Turing test was described by A. Turing in his paper [4] as follows: An interrogator questions both Turing machine and second participant of the test (a person), each of which tries to appear human. The interrogator does not know from whom exactly he receives answers and has the objective to tell the Turing machine from the person (for more details, see, for example, [3]).

In this paper, we consider a formalization of the Turing test and obtain the following results (Figure 1.1):

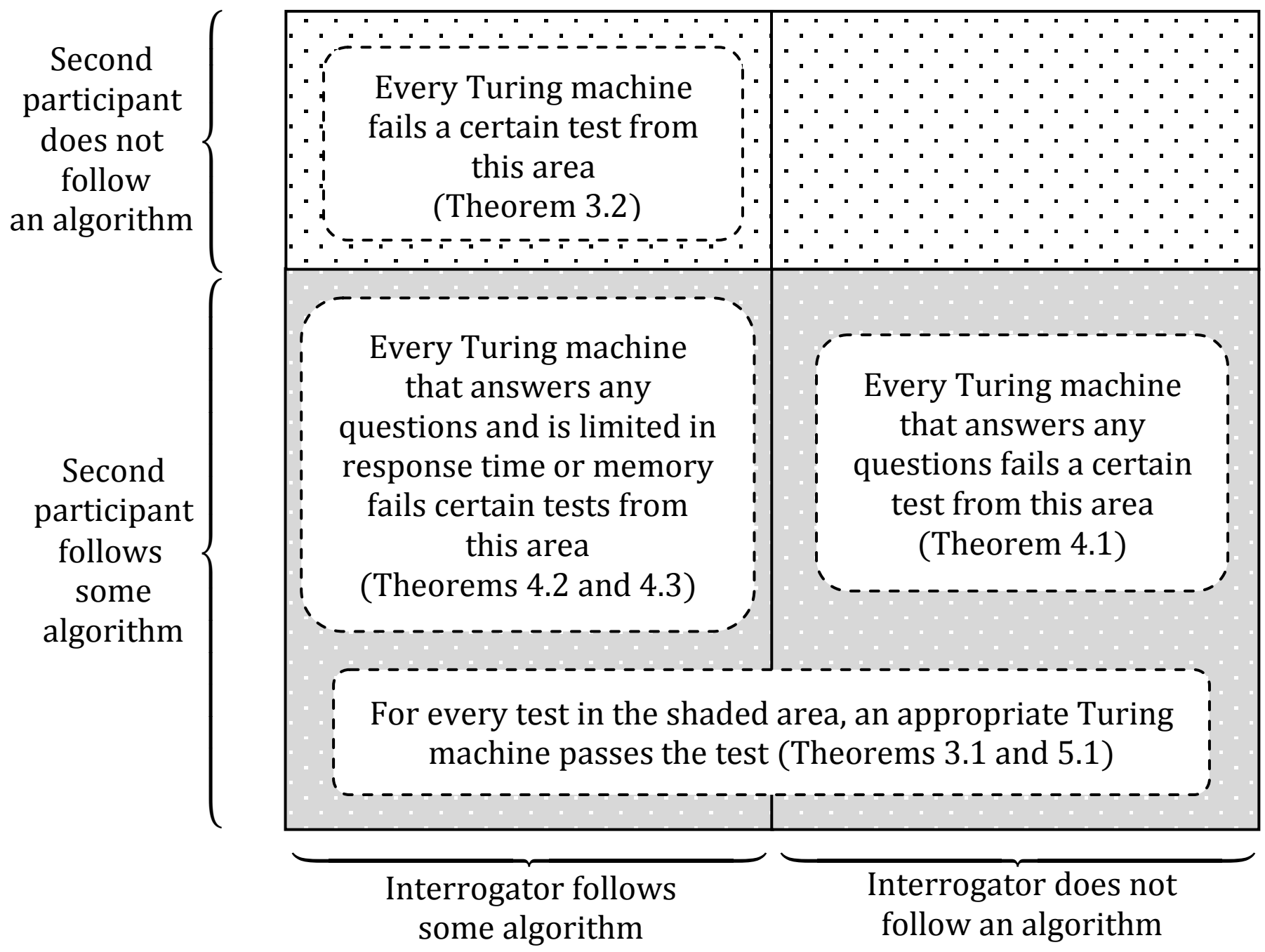

Figure 1.1. Summary of results

The rest of the paper is organized as follows:

Section 2 gives definitions and introduces notations.
Section 3 considers tests with arbitrary Turing machines.
Section 4 considers tests with Turing machines of some special classes.
Section 5 considers the strict Turing test.
Section 6 presents the conclusion.
Section 7 contains a list of definitions and notations.
Section 8 contains references.
Section 9 is an appendix, where we consider a random answers model for the second participant of the test that does not follow an algorithm.

## 2. General definitions and notations

### 2.1. <u>Words and numbers</u>

1. Suppose $\mathbb{A}$ is a finite alphabet and symbol $\theta$ does not belong to $\mathbb{A}$. By $\mathbb{A}^*$ denote all words over $\mathbb{A}$, including the empty word $\lambda$. By $\lambda^n$ denote the sequence of $n$ empty words $\lambda$.
2. Denote by $\mathbb{N}$ the set of all natural numbers (excluding zero), and put $\mathbb{N}_0 \overset{\text{def}}{=} \mathbb{N} \cup \{0\}$. We will not make any distinction between the elements of $\mathbb{N}_0$ and their notations over the alphabet $\mathbb{A}$, assuming $\mathbb{N}_0 = \mathbb{A}^* \backslash \{\lambda\}$.
3. Fix some letter $a \in \mathbb{A}$, and for each word $\omega \in \mathbb{A}^*$ denote by $\overline{\omega}$ the word $a\omega$.

4. Put $\mathbb{B} \overset{\text{def}}{=} \mathbb{A} \cup \{\theta\}$.

2.2. Turing machines

1. We assume that:
   - Turing machine (TM) starts on the work tape that contains a finite string of symbols, which may be blank.
   - TM has several final states; some of them may be marked with "Left" or "Right".
   - Along with conventional work head, TM has three additional heads and three tapes each of which has a leftmost cell but is infinite to the right.

We name these additional heads as the oracle interface, the input and the output of TM, considering that the oracle interface and the input can only read out symbols from $\mathbb{B}$, and the output can only print symbols from $\mathbb{B}$, while $\theta$ stands for the blank symbol.

The tape of the oracle interface is blank (we shall use it for oracle TM; see Item 2.4.1 below), and both input and output tapes are replaced with new tapes when TM starts up or when TM is transferred to the initial state.

2. Denote by $\mathbb{M}$ the set of all TMs that we have described. We shall not distinguish between TMs and their descriptions (in the form of the words over $\mathbb{A}$) suitable for realization on the universal TM.

3. Let us consider the functioning of TM as an exchange of questions and answers.

A question is sent to TM only when it is in the initial state or in the final state (in the latter case, TM is moved to the initial state, but the content of the work tape does not change). The question is a word over $\mathbb{B}$ recorded at the beginning of the new input tape. After installation of the input tape, TM starts up, and if TM halts, the maximum length word over $\mathbb{A}$, placed at the beginning of the output tape, is the answer to the question (thus the answer may be equal to $\lambda$).

4. We say that:
   - TM $\mathfrak{M}$ answers the questions $\beta_1, \dots, \beta_n$ if $\mathfrak{M}$ calculates those answers in a finite number of cycles. In the specified case denote by $\mathfrak{M}(\beta_1, \dots, \beta_n)$ the answer of $\mathfrak{M}$ to the last question $\beta_n$; denote the reverse case by the formal equation $\mathfrak{M}(\beta_1, \dots, \beta_n) = \infty$ [1].
   - TM $\mathfrak{M}$ recognizes TM $\mathfrak{N}$ if $\mathfrak{M}$ answers the question $\mathfrak{N}$.
   - TM $\mathfrak{M}$ is communicable if $\mathfrak{M}$ answers any sequence of questions, each of which belongs to $\mathbb{A}^*$.
   - TM $\mathfrak{M}$ is autonomous if the answers of $\mathfrak{M}$ do not depend on the content of the questions.
   - TM $\mathfrak{M}$ is the generator if $\mathfrak{M}$ is simultaneously communicable and autonomous.

   Denote by $\mathbb{C}$ the set of all communicable TMs. For generator $\mathfrak{H}$ and $n \in \mathbb{N}$ put $\mathfrak{H}[n] \overset{\text{def}}{=} \mathfrak{H}(\beta_1, \dots, \beta_n)$, where $\beta_1, \dots, \beta_n$ are some questions.

5. Assign to each TM $\mathfrak{M}$ the generator $\widetilde{\mathfrak{M}}$ that operates under the following principle: For each $n \in \mathbb{N}$, put $\widetilde{\mathfrak{M}}[n] \overset{\text{def}}{=} \begin{cases} \lambda, \text{ if } \mathfrak{M}(\lambda^n) = \infty, \\ \mathfrak{M}(\lambda^n) \text{ in another case.} \end{cases}$

6. We call TM $\mathcal{E}$ the enumerator if $\forall_{n \in \mathbb{N}} (\mathcal{E}(n) \in \mathbb{M})$.

7. For every enumerator $\mathcal{E}$, put $\mathcal{E}\{N\} \overset{\text{def}}{=} \{\mathcal{E}(n)\colon n = 1, \dots, N\}$. Fix the enumerator $\mathcal{A}$ that enumerates all TMs: $\mathcal{A}\{\infty\} = \mathbb{M}$. Put $\mathfrak{A}_k \overset{\text{def}}{=} \mathcal{A}(k)$.

8. Assign to every enumerator $\mathcal{E}$ that enumerates some communicable TMs $(\mathcal{E}\{\infty\} \subseteq \mathbb{C})$ the generator $\overline{\mathcal{E}}$ that is constructed by analogy with Cantor's diagonal method: $\overline{\mathcal{E}}[n] \overset{\text{def}}{=} \overline{\mathcal{E}(n)(\lambda^n)}$.

[1] Assuming $\infty \notin \mathbb{A}$.

2.3. Oracles

1. Denote by oracle the content of the oracle interface tape, the cells on which contain symbols from $\mathbb{B}$.

2. Denote by $\Pi$ the oracle that contains the lexicographic ordered notations $\mathfrak{M}\theta\mathfrak{N}\theta$ for all TMs $\mathfrak{M}$ and $\mathfrak{N}$ such that $\mathfrak{M}$ recognizes $\mathfrak{N}$.

3. Denote by $\Theta$ the blank oracle.

2.4. Oracle Turing machines

1. We shall understand the construction from Item 2.2.1, where the oracle interface tape may be non-blank, as oracle Turing machine (OTM). Denote OTM by $\mathfrak{M}^{\Phi}$, where $\mathfrak{M}$ is TM and $\Phi$ is oracle (thus $\mathfrak{M}^{\Theta} = \mathfrak{M}$).

2. Extending the concepts from Items 2.2.3 and 2.2.4 to OTMs, we call OTMs $U$ and $V$ as $N$-similar if for every $n \leq N$ and for any questions $\beta = \beta_1, \dots, \beta_n$ the equality $U(\beta) = V(\beta)$ holds. Two OTMs $U$ and $V$, which are $N$-similar for every $N \in \mathbb{N}$, we call similar and denote this by $U \approx V$.

3. We say that OTM $Q$ reduces to TM $\mathfrak{M}$ if $Q \approx \mathfrak{M}$. Denote the type of OTM by notation $\succ\mathbb{M}$ if OTM reduces to some TM and in another case use the notation $\nsucc\mathbb{M}$.

2.5. Testers

1. Any pair $\langle I, Q \rangle$, where $I$ and $Q$ are OTMs, we call the tester. For the specified tester we call $I$ the interrogator, and $Q$ the second participant (SP).

2. Denote the tester type by notation (type of interrogator, type of SP).

3. We say that interrogator is dumb if its answer to any consequence of questions is $\lambda$.

2.6. Tests

1. Consider tester $T = \langle I, Q \rangle$ and TM $\mathfrak{M}$ and define the following procedure as the left (the right) test:
   - At the beginning of the test, the question $\lambda$ enters the interrogator $I$ input.
   - Each answer of interrogator serves as the question for both $Q$ and $\mathfrak{M}$ (we call such a question the test question).
   - The question $\alpha_Q\theta\alpha_{\mathfrak{M}}$ (or the question $\alpha_{\mathfrak{M}}\theta\alpha_Q$ for the right test), where $\alpha_Q$ and $\alpha_{\mathfrak{M}}$ are the answers of $Q$ and $\mathfrak{M}$ to the last test question, enters $I$ input. We interpret this as receiving the first test answer from the "left" subject of the test, and receiving the second test answer from the "right" one.
   - Whenever $I$ has reached the final state marked with "Left" ("Right"), the test is finished and the result of the test is "SP is on the left" ("SP is on the right").

   Denote by $[T, \mathfrak{M}, \ell]$ (by $[T, \mathfrak{M}, r]$) the left (the right) test described above and denote by $[T, \mathfrak{M}, *]$ any of these tests.

2. We say that:
   - TM $\mathfrak{M}$ fails the left test $[\langle I, Q \rangle, \mathfrak{M}, \ell]$ (the right test $[\langle I, Q \rangle, \mathfrak{M}, r]$) if either $\mathfrak{M}$ does not answer some test question that SP $Q$ has answered or if this test is finished with the correct result.
   - TM fails the test $[T, \mathfrak{M}]$ if it fails both the left test $[T, \mathfrak{M}, \ell]$ and the right test $[T, \mathfrak{M}, r]$[2].

   The statement that TM $\mathfrak{M}$ fails the test $[T, \mathfrak{M}]$ we denote in short by $\mathfrak{M} \ntriangleright T$; the converse statement we denote by $\mathfrak{M} \triangleright T$.

3. We say that tester $T$ is successful for a given set $\mathbb{L} \subseteq \mathbb{M}$ if $\forall_{\mathfrak{M} \in \mathbb{L}} (\mathfrak{M} \ntriangleright T)$.

4. Assign to each SP $\mathfrak{Q} \in \mathbb{M}$ the dumb interrogator $\mathfrak{I}_{\mathfrak{Q}} \in \mathbb{M}$ who completes the test $\left[\langle \mathfrak{I}_{\mathfrak{Q}}{}^{\Phi}, \mathfrak{Q} \rangle, \mathfrak{M}, *\right]$ for any oracle $\Phi$ only in the case when the test answers from $\mathfrak{Q}$ and $\mathfrak{M}$ are different, and in this case $\mathfrak{I}_{\mathfrak{Q}}{}^{\Phi}$ calculates $\alpha = \mathfrak{Q}(\lambda^n)$, where $n$ is a number of the current test step, and indicates the location of SP as the source of reply $\alpha$. Note that $\mathfrak{I}_{\mathfrak{Q}}{}^{\Phi}$ performs operations that do not depend on $\Phi$.

   Put $T_{\mathfrak{Q}}^{\Phi} \stackrel{\text{def}}{=} \langle \mathfrak{I}_{\mathfrak{Q}}{}^{\Phi}, \mathfrak{Q} \rangle$ and $\mathbb{T}^{\Phi} \stackrel{\text{def}}{=} \left\{ T_{\mathfrak{Q}}^{\Phi} : \mathfrak{Q} \in \mathbb{M} \right\}$.

2.7. Comments

1. The handling of every question by certain TM with the input head and the output head can be described as the functioning of an arbitrary TM with only one head and with initial content of the work tape equal to the question (see [2]). Then note that in accordance with § 42 [1], there is an elementary arithmetic formula $\notin(x,y)$ with free occurrence of variables $x$ and $y$, which is true for $x \in \mathbb{M}$ and $y \in \mathbb{M}$ if and only if $x$ and $y$ are on the tape $\Pi$.

[2] See also the strict test in Section 5.

2. We consider SP as the tester constituent and explain this as follows.

The original description of the test offered by A. Turing assumes the loyalty of SP to interrogator. Now if interrogator gives in his zero numbered test question the instructions on how SP should operate (for example, instructions in the form $\mathfrak{M}\theta\notin$ for OTM $\mathfrak{M}^{\Pi}$), then SP, due to his loyalty, will fulfill these instructions. The instructions to simulate the OTM equipped with computable oracle are executable by a human. The problem of simulation of OTM equipped with non-computable oracle is beyond the scope of this paper.

## 3. Tests with arbitrary Turing machines

We shall consider tests under various conditions of reducibility of interrogator and SP to TM (Theorems 3.1 and 3.2) and prove at first the following lemma.

LEMMA 3.1.

1. *For each enumerator $\mathcal{E}$ that enumerates some communicable TMs*[3] *and for each oracle $\Phi$, the tester $T_{\overline{\mathcal{E}}}^{\Phi}$ is successful for $\mathcal{E}\{\infty\}$.*

2. *Tester $T_{\overline{\mathcal{E}}}^{\Phi}$ will finish the test $\tau = \left[T_{\overline{\mathcal{E}}}^{\Phi}, \mathcal{E}(n),*\right]$ not later than at the nth step.*

PROOF. The interrogator $\mathfrak{I}_{\overline{\mathcal{E}}}^{\Phi}$ is mute, and due to inequality $\overline{\mathcal{E}}[n] \neq \big(\mathcal{E}(n)\big)(\lambda^n)$, this interrogator successfully completes the test $\tau$ not later than at the $n$th step. □

THEOREM 3.1.

1. *For each tester with SP that reduces to TM, the specified TM can pass the test; hence, any such tester is not successful for* $\mathbb{M}$ [4].

2. *For each tester $T$ of the type* $(\succ\mathbb{M}, \succ\mathbb{M})$, *some generator $\mathfrak{H}$ can pass the test* $[T, \mathfrak{H}]$.

3. *The problem of constructing the generator $\mathfrak{H}$ that can pass the test* $[\langle\mathfrak{I}, \mathfrak{Q}\rangle, \mathfrak{H}]$, *where $\mathfrak{I}$ and $\mathfrak{Q}$ are arbitrary TMs, is algorithmically unsolvable.*

PROOF. The proof of Item 1 of the theorem arises from the definition of TM's ability to pass the test.

To prove Item 2 of the theorem consider $\mathfrak{H} = \widetilde{\mathfrak{M}}$, where TM $\mathfrak{M}$ is defined as follows: If $\mathfrak{I}$ and $\mathfrak{Q}$ are TMs, to which interrogator and SP are reduced, then $\mathfrak{M}$ is an autonomous TM that simulates $[\langle\mathfrak{I}, \mathfrak{Q}\rangle, \mathfrak{Q},*]$ and gives the answers that are the same as those of $\mathfrak{Q}$ in this test.

In order to prove Item 3 of the theorem assume that there exists TM $\mathfrak{B}$ such that generator $\mathfrak{H} = \mathfrak{B}(\mathfrak{I}\theta\mathfrak{Q})$ can pass the test $[\langle\mathfrak{I}, \mathfrak{Q}\rangle, \mathfrak{H}]$. Then consider the following enumerator $\mathcal{E}$ that enumerates generators: $\mathcal{E}(n) \overset{\text{def}}{=} \mathfrak{B}\big(\mathfrak{I}_{\mathfrak{A}_n}\theta\mathfrak{A}_n\big)$. According to the definition of $\mathcal{E}$, we have $\forall_{n\in\mathbb{N}}\big(\mathcal{E}(n) \rhd T_{\mathfrak{A}_n}^{\Theta}\big)$ and thus $\forall_{T\in\mathbb{T}^{\Theta}}\exists_n(\mathcal{E}(n) \rhd T)$. But Item 1 of Lemma 3.1 implies that $\exists_{T\in\mathbb{T}^{\Theta}}\forall_n(\mathcal{E}(n) \not\rhd T)$. □

THEOREM 3.2. *Some tester of the type* $(\succ\mathbb{M}, \not\succ\mathbb{M})$ *with dumb interrogator is successful for* $\mathbb{M}$.

PROOF[5]. Let us describe the desired tester $T = \langle I, Q\rangle$, where $Q$ is equipped with oracle $\Pi$:

SP $Q$, after receiving the $n$th test question, acts as follows:

- $Q$ calculates $\mathfrak{A}_n$ and queries its oracle $\Pi$ whether $\mathfrak{A}_n$ recognizes itself.
- If the oracle answers in the affirmative, $Q$ simulates the work of $\mathfrak{A}_n$ with question $\mathfrak{A}_n$, calculates the number $t$ of cycles that is necessary for $\mathfrak{A}_n$ to compute $\mathfrak{A}_n(\mathfrak{A}_n)$, and gives the answer $t$.

---

[3] Note that if there is an enumerator $\mathcal{E}$ that enumerates all communicable TMs, then $\mathcal{E}\{\infty\}$ would include all generators, and then any generator would be similar to some element in $\mathcal{R}\{\infty\}$, where $\mathcal{R}(n) \overset{\text{def}}{=} \widetilde{\mathcal{E}(n)}$, but $\overline{\mathcal{R}} \notin \mathcal{R}\{\infty\}$.

[4] Thus, for the successfulness of such tester either SP should not reduce to TM (see Theorem 3.2) or SP, which reduces to TM, should not belong to the class of TMs that are subjects of the test (see Section 4).

[5] The idea of the proof is based on Subsection 3 of Section 6 of [4].

- If the oracle answers negative, $Q$ gives the answer 0.

Interrogator $I$, after receiving the answers to the $n$th test question, follows the next procedure:

- If the answers are equal, $I$ continues the test.
- Otherwise, $I$ treats each nonzero answer as a notation of some natural number $t$, calculates $\mathfrak{A}_n$ and verifies, whether $\mathfrak{A}_n$ will answer the question $\mathfrak{A}_n$ for $t$ cycles; at negative result of verification $I$ finishes the test, locating SP as the source of another answer.

If some TM $\mathfrak{N}$ passes the test $[T, \mathfrak{N}]$, then the problem of recognition of TMs not recognizing themselves is solvable by $\mathfrak{N}$, but it is well known that this is impossible (see, for example, § 42 [1]). To complete the proof, note that reducibility of $Q$ to TM contradicts Item 1 of Theorem 3.1. □

## 4. Testing for special classes of Turing machines

In this section, we consider the tests for all communicable TMs, and the tests for communicable TMs with time or memory limitation (we can suppose that these TMs are close-to-reality models of computer programs).

### 4.1. Testing for communicable Turing machines

THEOREM 4.1.

1. *Any tester of the type* $(\succ\mathbb{M}, \succ\mathbb{M})$ *is not successful for* $\mathbb{C}$.
2. *Some tester of the type* $(\succ\mathbb{M}, \nsucc\mathbb{M})$ *is successful for* $\mathbb{C}$.
3. *Some tester of the type* $(\nsucc\mathbb{M}, \succ\mathbb{M})$ *is successful for* $\mathbb{C}$.

PROOF. Items 1 and 2 of the theorem follow from Item 2 of Theorem 3.1 and from Theorem 3.2, respectively. To prove Item 3 of the theorem we shall describe the desired interrogator $I$, which is equipped with oracle $\Pi$, and SP $\mathfrak{Q} \in \mathbb{M}$.

During the test $I$ makes use of parameter $r \in \mathbb{N}_0$, supposing $r = 0$ at the beginning of the test. At the $n$th step of the test ($n \in \mathbb{N}$) interrogator, by means of oracle $\Pi$, finds the minimal $k_n > r$, at which:

- If $n = 1$, then $\mathfrak{A}_{k_n}$ recognizes itself.
- If $n > 1$, then $\mathfrak{A}_{k_n}$ answers the questions $\mu_1, \mu_2, \dots, \mu_{n-1}, \mathfrak{A}_{k_n}$, where $\mu_1, \mu_2, \dots, \mu_{n-1}$ are the previous test questions (each of which belongs to $\mathbb{M}$).

Thereafter $I$ assigns the value $k_n$ to $r$ and puts the $n$th test question $\mu_n \overset{\text{def}}{=} \mathfrak{A}_{k_n}$ [6].

SP $\mathfrak{Q}$ gives the answer $\overline{\mu_n(\mu_1, \mu_2, \dots, \mu_n)}$ to the $n$th test question. For the first time when $I$ received two different answers, $I$ calculates the answer of $\mathfrak{Q}$ and completes the test, locating $\mathfrak{Q}$ as the source of its answer.

Now consider the test $[\langle I, \mathfrak{Q}\rangle, \mathfrak{C}]$ for arbitrary $\mathfrak{C} \in \mathbb{C}$. The proof of the theorem follows from $\mathfrak{C} = \mathfrak{A}_{k_n}$ for some $n$, while the reducibility of $I$ to TM contradicts Item 1 of this theorem, which is already proved[7]. □

### 4.2. Testing for Turing machines limited in time

We say that communicable TM $\mathfrak{C}$ is limited in time if there is an upper bound for number of cycles required to calculate each following answer of $\widetilde{\mathfrak{C}}$.

THEOREM 4.2. *Some tester of the type* $(\succ\mathbb{M}, \succ\mathbb{M})$ *with dumb interrogator is successful for all communicable limited in time TMs.*

PROOF. For each TM $\mathfrak{M}$ and each $t \in \mathbb{N}$ denote by $\mathfrak{M}|_t$ TM $\mathfrak{M}$ that operates under the control of special supervisor TM. The supervisor observes how $\mathfrak{M}$ processes with questions $\lambda$, and at the first time when $\mathfrak{M}$ has not provide an answer for $t$ work cycles, supervisor shuts down $\mathfrak{M}$ and gives $\lambda$ as the answer of $\mathfrak{M}|_t$ to the current and all subsequent questions.

---

[6] Note that the interrogator's questions do not depend on the receiving answers.

[7] Note that Item 1 of Theorem 3.1 implies that $\mathfrak{Q} \notin \mathbb{C}$, and this is also clear from the principle of how $\mathfrak{Q}$ works: If $\mathfrak{Q} \in \mathbb{C}$, then $\mathfrak{Q}(\mu_1, \mu_2, \dots, \mu_n) = \overline{\mathfrak{Q}(\mu_1, \mu_2, \dots, \mu_n)}$ for some $n$.

Now construct an enumerator $\mathcal{E}$ that enumerates $\mathfrak{M}|_t$ for all $\mathfrak{M} \in \mathbb{M}$ and all $t \in \mathbb{N}$. For each $\mathfrak{C} \in \mathbb{C}$ that is limited in time, an arbitrary TM in $\mathcal{E}\{\infty\}$ is similar to $\widetilde{\mathfrak{C}}$, whereby the theorem statement follows from Item 1 of Lemma 3.1 [8]. □

4.3. Testing for Turing machines limited in memory

Consider a class of communicable TMs limited in memory, namely, TMs with a uniform upper bound for the number of states and a uniform upper bound for the length of the work tape segment that TM may scan in the processing of empty questions.

THEOREM 4.3. *Some tester of the type* $(\succ\mathbb{M}, \succ\mathbb{M})$ *with dumb interrogator is successful for all communicable limited in memory TMs.*

PROOF. We need the following definitions:

- Denote by $\mathbb{L}$ the set of all TMs that satisfy the specified limitation for the number of states and for the length of the initial content of the work tape.
- Denote by $\mathbb{K}$ the set of all TMs from $\mathbb{L} \cap \mathbb{C}$ that satisfy the specified limitation for the length of the work tape segment that TM makes use of in the processing of empty questions.

Now calculate such $N$ that $\mathbb{L} \subseteq \mathcal{A}\{N\}$ and put $\mathbb{L}_N \stackrel{\text{def}}{=} \{\mathfrak{M} \in \mathbb{L}: \mathfrak{M}(\lambda^N) \neq \infty\}$ (note that $\mathbb{K} \subseteq \mathbb{L}_N$). Observing the work of $\mathfrak{M} \in \mathbb{L}$ that calculates $\mathfrak{M}(\lambda^N)$, it is possible to reject those $\mathfrak{M}$ for which:

- The length of the work tape segment that $\mathfrak{M}$ makes use of indicates that $\mathfrak{M} \notin \mathbb{K}$.
- At the processing of current question λ, the configuration of $\mathfrak{M}$ (i.e. the combination (state of $\mathfrak{M}$, position and content of the work tape)) was repeated, whence it follows that $\mathfrak{M} \notin \mathbb{L}_N$.

Thus we can reject all TMs from $\mathbb{L}\backslash\mathbb{L}_N$ and some TMs from $\mathbb{L}_N\backslash\mathbb{K}$. Then it is possible to construct an enumerator $\mathcal{R}$, based on $\mathcal{A}$, on the definition of $\mathbb{L}$, and on the described above rejection, such that $\mathbb{K} \subseteq \mathcal{R}\{N\} \subseteq \mathbb{L}_N$. Finally, it is possible to construct the following enumerator $\mathcal{E}$ that enumerates generators:

If $n \leq N$, then

$$\mathcal{E}(n)[k] \stackrel{\text{def}}{=} \begin{cases} \mathcal{R}(n)(\lambda^k) \text{ if } k \leq N, \\ \lambda \text{ if } k > N; \end{cases}$$

and if $n > N$, then $\forall_{k \in \mathbb{N}}(\mathcal{E}(n)[k] \stackrel{\text{def}}{=} \lambda)$.

For each $\mathfrak{C} \in \mathbb{K}$ there is an arbitrary TM in $\mathcal{E}\{N\}$ that is $N$-similar to $\widetilde{\mathfrak{C}}$, whereby the theorem statement follows from Item 2 of Lemma 3.1 [9]. □

## 5. The strict Turing Test

5.1. Definition of the strict Turing Test

TM that is the subject of the test has some advantages: If this TM does not answer the test question simultaneously with SP, it passes the test (see Item 2.6.2). The choice of this condition is rather arbitrary and that allows us to consider the "strict" Turing test:

- TM fails the left (the right) strict test if either TM does not answer some test question (regardless whether SP has answered this question) or this test is finished with the correct result.
- TM fails the strict test if it fails both the left and the right strict test.

We introduce for the strict tests notations $[\![T, \mathfrak{M}, l]\!]$, $[\![T, \mathfrak{M}, r]\!]$, $[\![T, \mathfrak{M}, *]\!]$, $[\![T, \mathfrak{M}]\!]$, $\mathfrak{M} \trianglerighteq T$, and $\mathfrak{M} \ntrianglerighteq T$ by analogy with the notations $[T, \mathfrak{M}, l]$, $[T, \mathfrak{M}, r]$, $[T, \mathfrak{M}, *]$, $[T, \mathfrak{M}]$, $\mathfrak{M} \vartriangleright T$, and $\mathfrak{M} \ntriangleright T$ for the ordinary tests.

It is clear that if $\mathfrak{M} \ntriangleright T$, then $\mathfrak{M} \ntrianglerighteq T$. Therefore, the statements of Lemma 3.1, Theorems 3.2, 4.2, 4.3, and Items 2 and 3 of Theorem 4.1 are extended to the case of the strict test. A different situation arises with the statement of Theorem 3.1 (and its corollary, Item 1 of Theorem 4.1): If SP of the tester $T$ reduces to TM $\mathfrak{Q}$, then $\mathfrak{Q}$ may fail the strict test $[\![T, \mathfrak{Q}]\!]$ (take for example TM $\mathfrak{Q}$ that does not answer any question). However, the modified Theorem 3.1, given in following subsection, is valid for the strict test.

---

[8] Note that Item 1 of Theorem 3.1 implies that SP of the described tester is not limited in time, and this is also clear from the principle of how this SP works.

[9] Note that from Item 1 of Theorem 3.1 it follows that SP of the described tester does not belong to $\mathbb{K}$, and this is also clear from the principle of how this SP works.

5.2. Successfulness of the testers for the strict Turing test

THEOREM 5.1.

1. *For each tester, which SP is reducible to TM, an arbitrary TM can pass the strict test with this tester*[10].

2. *For each tester $T$ of the type* $(\succ\mathbb{M}, \succ\mathbb{M})$ *some generator* $\mathfrak{H}$ *can pass the strict test* $[\![T, \mathfrak{H}]\!]$[11].

3. *For each tester $T$ of the type* $(\succ\mathbb{M}, \succ\mathbb{M})$ *some communicable TM* $\mathfrak{C}$ *can pass the strict test* $[\![T, \mathfrak{C}]\!]$[12].

4. *For a given oracle $\Phi$ and for any TMs $\mathfrak{I}$ and $\mathfrak{Q}$, the problem of constructing the TM $\mathfrak{M}$ that can pass the strict test $[\![\langle\mathfrak{I}^{\Phi}, \mathfrak{Q}\rangle, \mathfrak{M}]\!]$ is algorithmically unsolvable*[13].

PROOF OF ITEMS 1, 2, AND 3 OF THEOREM 5.1.

For an arbitrary tester $T$, denote by $\mathfrak{Q}$ the TM to which SP of this tester is reduced, and consider the strict test $\tau = [\![T, \mathfrak{Q}, *]\!]$. If $\mathfrak{Q}$ answers all test questions, then the proof of Items 1 and 2 of the theorem coincides with the proof of Items 1 and 2 of Theorem 3.1. Assume now that $\mathfrak{Q}$ does not answer the $n$th test question in the test $\tau$, where $n \in \mathbb{N}$. Then $\mathfrak{Q}$ fails the test $\tau$, but generator $\mathfrak{H}$, which first $n-1$ answers are equal to those of $\mathfrak{Q}$ in the test and the following answers are equal to $\lambda$, will pass the test $[\![T, \mathfrak{H}, *]\!]$. That completes the proof of Items 1 and 2 of Theorem 5.1.

Finally, generator is a communicable TM and hence Item 3 of Theorem 5.1 is the corollary of Item 2 of Theorem 5.1. □

5.3. Lemmas

To prove Item 4 of Theorem 5.1 let us:

- Fix the oracle $\Phi$ from the hypothesis of this item.
- Assume the existence of TM $\mathfrak{F}$ that was specified in Item 4 of the theorem: $\mathfrak{F}(\mathfrak{I}\theta\mathfrak{Q}) \trianglerighteq \langle\mathfrak{I}^{\Phi}, \mathfrak{Q}\rangle$.
- Prove the following lemmas.

LEMMA 5.1. *For each SP $\mathfrak{Q}$ and each TM $\mathfrak{M}$, the following values are equal for $[\![T_{\mathfrak{Q}}^{\Phi}, \mathfrak{M}, \ell]\!]$ and $[\![T_{\mathfrak{Q}}^{\Phi}, \mathfrak{M}, r]\!]$:*

1. *The numbers of tests' steps.*
2. *The numbers (maybe infinite) and contents of answers of $\mathfrak{Q}$.*
3. *The numbers (maybe infinite) and contents of answers of $\mathfrak{M}$.*

PROOF. By definition of interrogator ${\mathfrak{I}_{\mathfrak{Q}}}^{\Phi}$, it treats the answers of the subjects of the test symmetrically, and that proves Item 1 of the lemma. Items 2 and 3 of the lemma follow from Item 1 of this lemma and from the dumbness of ${\mathfrak{I}_{\mathfrak{Q}}}^{\Phi}$. □

Taking Lemma 5.1 into account for the strict tests with a tester that has a form of $T_{\mathfrak{Q}}^{\Phi}$, we shall discuss the number of the steps in the test and the answers of the subjects of the test without specifying the orientation of the test (the left one or the right one).

To formulate and prove Lemmas 5.2, 5.3, 5.4, and 5.5 we need the following definitions and notations.

- Denote by $|\nu|$ the number of words in the sequence $\nu$ of words over $\mathbb{A}$; we say that sequences of words $\eta$ and $\mu$ satisfy the relation $\eta \dashv \mu$ if any of the following cases holds:
  - $0 = |\eta| < |\mu|$.
  - $\infty \neq |\eta| < |\mu|$ and $\eta$ is the beginning of $\mu$.
  - $|\eta| = |\mu| = \infty$ and $\eta = \mu$.

---

[10] An analogue of Item 1 of Theorem 3.1.

[11] An analogue of Item 2 of Theorem 3.1.

[12] An analogue of Item 1 of Theorem 4.1.

[13] That accentuates the difference between the strict and the ordinary tests: See Item 1 of Theorem 3.1.

- Fix some SP $\mathfrak{Q}$ and put $\mathfrak{Q}_0 \stackrel{\text{def}}{=} \mathfrak{Q}$, $\mathfrak{Q}_{n+1} \stackrel{\text{def}}{=} \mathfrak{F}(\mathfrak{I}_{\mathfrak{Q}_n}\theta\mathfrak{Q}_n)$, $I_n \stackrel{\text{def}}{=} \mathfrak{I}_{\mathfrak{Q}_n}{}^{\Phi}$, $T_n \stackrel{\text{def}}{=} T^{\Phi}_{\mathfrak{Q}_n}$, where $n \in \mathbb{N}_0$.
- For $n > 0$ and the strict test $\llbracket T_{n-1}, \mathfrak{Q}_n \rrbracket$:
  - Denote by $\chi^{(n)}$ the sequence of answers of SP $\mathfrak{Q}_{n-1}$ in this test.
  - Denote by $\gamma^{(n)} = \gamma^{(n)}_1, \gamma^{(n)}_2, \ldots$ the sequence of answers of TM $\mathfrak{Q}_n$ that is the subject of this test.

LEMMA 5.2. $\forall_{n\in\mathbb{N}}(\chi^{(n)} \dashv \gamma^{(n)})$.

PROOF. For $n > 0$, $\mathfrak{Q}_n \trianglerighteq T_{n-1}$. Thus, owing to the specificity of how interrogator $I_n$ operates, only the two following cases are possible:

1. The number of test steps is infinite, and moreover, $\chi^{(n)} = \gamma^{(n)}$.
2. SP $\mathfrak{Q}_{n-1}$ has not answered to the $j$th ($j \geq 1$) test question that has been answered by TM $\mathfrak{Q}_n$. Thus $|\gamma^{(n)}| = |\chi^{(n)}| + 1$ and this implies that for $j = 1$, the equation $|\chi^{(n)}| = 0$ is satisfied, and for $j > 1$, $\chi^{(n)} = \gamma^{(n)}_1, \gamma^{(n)}_2, \ldots, \gamma^{(n)}_{j-1}$. □

LEMMA 5.3. $\forall_{n\in\mathbb{N}}(\gamma^{(n)} \dashv \gamma^{(n+1)})$.

PROOF. Consider the tests $\llbracket T_{n-1}, \mathfrak{Q}_n, * \rrbracket$ and $\llbracket T_n, \mathfrak{Q}_{n+1}, * \rrbracket$. The interrogators $I_{n-1}$ and $I_n$ are dump, and $\mathfrak{Q}_{n+1} \trianglerighteq T_n$. Thus $\gamma^{(n)} = \chi^{(n+1)}$ or $\gamma^{(n)} \dashv \chi^{(n+1)}$, due to the definition of $I_n$. Then the proof follows from Lemma 5.2. □

LEMMA 5.4. $\forall_{n\in\mathbb{N}}(|\gamma^{(n)}| \geq n)$.

PROOF. Lemma 5.2 implies that $|\gamma^{(1)}| > 0$, and for $n > 1$ the statement of the lemma follows from Lemma 5.3. □

LEMMA 5.5. *For each* $n \in \mathbb{N}$ *and for every* $k$, *where* $0 < k \leq |\gamma^{(n)}|$, *the word* $\gamma^{(n)}_k$ *is calculated algorithmically.*

PROOF. Owing to definition of $\mathfrak{I}_{\mathfrak{M}}$, this TM can be algorithmically constructed from TM $\mathfrak{M}$. Then, due to definition of $\mathfrak{Q}_n$, every $\mathfrak{I}_{\mathfrak{Q}_n}$ and thus every $\mathfrak{Q}_n$ can be algorithmically constructed from TM $\mathfrak{Q}$. Finally, $\gamma^{(n)}_k = \mathfrak{Q}_n(\lambda^k)$. □

5.4. Proof of Item 4 of Theorem 5.1

To prove Item 4 of Theorem 5.1 assign to each $\mathfrak{Q} \in \mathbb{M}$ the following generator $\mathfrak{H}_{\mathfrak{Q}}$: $\mathfrak{H}_{\mathfrak{Q}}[n] \stackrel{\text{def}}{=} \gamma^{(n)}_n$ ($\mathfrak{H}_{\mathfrak{Q}}$ is well defined due to Lemmas 5.4 and 5.5). Now, if $\gamma$ is the sequence of all answers of $\mathfrak{H}_{\mathfrak{Q}}$, then according to Lemmas 5.3 and 5.4, $\gamma^{(1)} \dashv \gamma$, hence $(\mathfrak{Q}_1 \trianglerighteq T^{\Phi}_{\mathfrak{Q}}) \Rightarrow (\mathfrak{H}_{\mathfrak{Q}} \trianglerighteq T^{\Phi}_{\mathfrak{Q}}) \Rightarrow (\mathfrak{H}_{\mathfrak{Q}} \triangleright T^{\Phi}_{\mathfrak{Q}})$. Now construct the following enumerator $\mathcal{E}$ that enumerates generators: $\mathcal{E}(n) \stackrel{\text{def}}{=} \mathfrak{H}_{\mathfrak{A}_n}$. We have $\forall_{n\in\mathbb{N}}(\mathcal{E}(n) \triangleright T^{\Phi}_{\mathfrak{A}_n})$ and thus $\forall_{T\in\mathbb{T}^{\Phi}}\exists_n(\mathcal{E}(n) \triangleright T)$. However, according to Item 1 of Lemma 3.1, $\exists_{T\in\mathbb{T}^{\Phi}}\forall_n(\mathcal{E}(n) \ntriangleright T)$. This contradiction proves that $\mathfrak{F}$ does not exist. □

# 6. Conclusion

The summary of results is given in Figure 1.1 (see Section 1). Note that some results can be extended to the case when each participant of the test can be an oracle Turing machine.

# 7. Definitions and notations

In tables given below, the global definitions and notations (terms) are shown with specifying the number of the subsection, where the corresponding definition or notation was denoted.

7.1. Definitions and notations

| Sets | | Symbols, words, and related concepts | | Special TMs | | Testers, interrogators, tests | |
|---|---|---|---|---|---|---|---|
| $\mathbb{A}, \mathbb{A}^*$ | 2.1.1 | $\theta$ | 2.1.1 | $\widetilde{\mathfrak{M}}$ | 2.2.5 | $\langle I, Q \rangle$ | 2.5.1 |
| $\mathbb{B}$ | 2.1.4 | $\lambda$ | 2.1.1 | $\mathfrak{A}_k$ | 2.2.7 | $T^{\Phi}_{\mathfrak{Q}}$ | 2.6.4 |

| | |
|---|---|
| $\mathbb{N}, \mathbb{N}_0$ | 2.1.2 |
| $\mathbb{M}$ | 2.2.2 |
| $\mathbb{C}$ | 2.2.4 |
| $\mathbb{T}^{\Phi}$ | 2.6.4 |

Relations

| | |
|---|---|
| $\approx$ | 2.4.3 |
| $\succ, \nsucc$ | 2.4.3 |
| $\rhd, \ntriangleright$ | 2.6.2 |
| $\trianglerighteq, \ntrianglerighteq$ | 5.1 |
| ⤙ | 5.3 |
| $\lambda^n$ | 2.1.1 |
| $\overline{\omega}$ | 2.1.3 |
| $\chi^{(n)}$ | 5.3 |
| $\gamma^{(n)}, \gamma_k^{(n)}$ | 5.3 |
| $\lvert\nu\rvert$ | 5.3 |

Oracles, OTM

| | |
|---|---|
| $\Pi$ | 2.3.2 |
| $\Theta$ | 2.3.3 |
| $\mathfrak{M}^{\Phi}$ | 2.4.1 |
| $\mathcal{E}\{N\}$ $(\mathcal{E}\{\infty\})$ | 2.2.7 |
| $\mathcal{A}$ | 2.2.7 |
| $\overline{\mathcal{E}}$ | 2.2.8 |
| $\mathfrak{Q}_n$ | 5.3 |
| $\mathfrak{F}$ | 5.3 |

Formula

| | |
|---|---|
| $\mathscr{f}$ | 2.7.1 |
| $T_n$ | 5.3 |
| $\mathfrak{I}_{\mathfrak{Q}}$ | 2.6.4 |
| $I_n$ | 5.3 |
| $[T, \mathfrak{M}, \ell], [T, \mathfrak{M}, r], [T, \mathfrak{M}, *]$ | 2.6.1 |
| $[T, \mathfrak{M}]$ | 2.6.2 |
| $[\![T, \mathfrak{M}, l]\!], [\![T, \mathfrak{M}, r]\!], [\![T, \mathfrak{M}, *]\!], [\![T, \mathfrak{M}]\!]$ | 5.1 |

Answers of (O)TM to questions

| | |
|---|---|
| $V(\beta), V(\beta) = \infty$ | 2.2.4, 2.4.2 |
| $\mathfrak{H}[n]$ | 2.2.4 |

## 7.2. Terms

| | | | | | |
|---|---|---|---|---|---|
| answer of (O)TM | 2.2.3, 2.4.2 | left (right) test | 2.6.1 | reducible to TM | 2.4.3 |
| answers to the questions, that, (O)TM | 2.2.4, 2.4.2 | limited in memory, TM | 4.3 | similar OTMs | 2.4.2 |
| autonomous TM | 2.2.4 | limited in time, TM | 4.2 | SP | 2.5.1 |
| communicable TM | 2.2.4 | $N$-similar OTMs | 2.4.2 | strict test | 5.1 |
| dumb interrogator | 2.5.3 | oracle | 2.3.1 | successful tester | 2.6.3 |
| enumerator | 2.2.6 | oracle interface of (O)TM | 2.2.1, 2.4.1 | test question | 2.6.1 |
| fails the test, that, TM | 2.6.2 | OTM | 2.4.1 | tester | 2.5.1 |
| generator | 2.2.4 | output of (O)TM | 2.2.1, 2.4.1 | TM | 2.2.1 |
| input of (O)TM | 2.2.1, 2.4.1 | question to (O)TM | 2.2.3, 2.4.2 | type of OTM | 2.4.3 |
| interrogator | 2.5.1 | recognition of TM | 2.2.4 | type of tester | 2.5.2 |

# 8. References

[1] Kleene S.C. (1967), 'Mathematical Logic', John Wiley and Sons, New York – London – Sydney.
[2] Papadimitriou, Christos H. (1994), 'Computational Complexity', Addison-Wesley.
[3] Saygin, A.P., Cicekli, I. and Akman, V. (2000), 'Turing Test: 50 Years Later', *Minds and Machines* 10, pp. 463-518.
[4] Turing, A. (1950), 'Computing Machinery and Intelligence', *Mind* 59(236), pp. 433–460.

# 9. Appendix. Probabilistic test

Consider some variation of Theorem 3.2: Let SP be equipped with the oracle that creates physically, namely, can be replaced with a random number generator.

There are at least two symbols in the alphabet $\mathbb{B}$: $\theta$ and $\overline{\lambda}$, and in order to use the notations that are conventional to binary random number generators we replace these symbols with 0 and 1. Then consider the following OTM $Z = \mathfrak{Z}^{\Xi}$:

- Oracle $\Xi$ is filled up with symbols 0 and 1 randomly and independently; where 0 appears with the probability $p_0$ and 1 appears with the probability $p_1$ (hence $\Xi$ is randomly chosen from the set of all oracles).
- TM $\mathfrak{Z}$ satisfies the following condition: If $\Xi = \xi_1 \xi_2 \ldots$, then $Z(\lambda^n) \overset{\text{def}}{=} \xi_n$.

THEOREM 9.1. *For some dumb interrogator* $\mathfrak{I}$ *of the type* $\succ\mathbb{M}$, *which depends on* $m \in \mathbb{N}$, *if* $p = \max(p_0, p_1) < 1$, *then* $\forall_{\mathfrak{C} \in \mathbb{C}} \left( P\{\mathfrak{C} \ntriangleright \langle \mathfrak{I}, Z \rangle\} \geq 1 - \frac{p^m}{1-p} \right)$.

PROOF. Without loss of generality, we equate each nonzero answer of any TM to 1. Now fix $m \in \mathbb{N}$ and construct the dumb interrogator $\mathfrak{I}$ by the following algorithm.

Interrogator $\mathfrak{I}$ consists of supervisor $\mathfrak{S}$ and two assistants: The left assistant $\mathfrak{L}$ and the right assistant $\mathfrak{R}$. The left (the right) assistant processes the answers of the left (the right) subject of the test and transmits the results of the processing to $\mathfrak{S}$.

After the $n$th answer has been received, an assistant carries out the following instructions:

1. Save the answer in memory and put $\delta = 0$.
2. For $k = 1, \dots, n$, start up $\mathfrak{A}_k(\lambda^n)$ for $n$ cycles of calculation.
3. If $\mathfrak{A}_k$ has provided not less than $m + k - 1$ answers, and these answers are equal to the received answers, put $\delta = 1$.
4. Report the value of $\delta$ to $\mathfrak{S}$.

Supervisor $\mathfrak{S}$ treats the messages $\delta_{\mathfrak{L}}$ and $\delta_{\mathfrak{R}}$ of $\mathfrak{L}$ and $\mathfrak{R}$ according to the following table.

| $\delta_{\mathfrak{R}}=$ / $\delta_{\mathfrak{L}}=$ | 0 | 1 |
|---|---|---|
| 0 | Continue the test | Complete the test with result "SP is on the left" |
| 1 | Complete the test with result "SP is on the right" | |

If some communicable TM $\mathfrak{C}$ has passed the test, then symbol 1 appears in the sequence of messages from the assistant that processes the answers of $Z$, and then for some $k$ the first $m + k - 1$ answers of $Z$ are equal to the answers of $\mathfrak{A}_k$. As a result, $P\{\mathfrak{C} \rhd \langle \mathfrak{I}, Z \rangle\} \leq \sum_{k=1}^{\infty} p^{m+k-1}$. □